\pdfoutput=1
\def\year{2020}\relax
\documentclass[letterpaper]{article} % DO NOT CHANGE THIS
\usepackage{aaai20}  % DO NOT CHANGE THIS
\usepackage{times}  % DO NOT CHANGE THIS
\usepackage{helvet} % DO NOT CHANGE THIS
\usepackage{courier}  % DO NOT CHANGE THIS
\usepackage[hyphens]{url}  % DO NOT CHANGE THIS
\usepackage{graphicx} % DO NOT CHANGE THIS
\usepackage{graphicx}  % DO NOT CHANGE THIS
\usepackage{booktabs}
\usepackage{adjustbox}
\usepackage{amsmath}
\usepackage{kotex}
\usepackage{ctable}
\usepackage{amsfonts}
\usepackage{bbm}
\usepackage{capt-of}
\usepackage{makecell} 
\usepackage[strings]{underscore}

\newcommand{\ie}{\textit{i}.\textit{e}., }
\newcommand{\eg}{\textit{e}.\textit{g}., }

\title{Deep Trustworthy Knowledge Tracing}
\author{Heonseok Ha, Uiwon Hwang, Yongjun Hong, Jahee Jang, and Sungroh Yoon\thanks{To whom correspondence should be addressed.}  \\
  Electrical and Computer Engineering\\
  Seoul National University\\
  Seoul 08826, Korea \\
  \{heonseok.ha, uiwon.hwang, yjhong, hukla, sryoon\}@snu.ac.kr \\
}

\begin{document}

\maketitle

\begin{abstract}
Knowledge tracing (KT), a key component of an intelligent tutoring system, is a machine learning technique that estimates the mastery level of a student based on his/her past performance. 
The objective of KT is to predict a student's response to the next question.
Compared with traditional KT models, deep learning-based KT (DLKT) models show better predictive performance because of the representation power of deep neural networks. 
Various methods have been proposed to improve the performance of DLKT, but few studies have been conducted on the reliability of DLKT.
In this work, we claim that the existing DLKTs are not reliable in real education environments. 
To substantiate the claim, we show limitations of DLKT from various perspectives such as knowledge state update failure, catastrophic forgetting, and non-interpretability.
We then propose a novel regularization to address these problems. The proposed method allows us to achieve trustworthy DLKT. In addition, the proposed model which is trained on scenarios with forgetting can also be easily extended to scenarios without forgetting.
\end{abstract}

\section{Introduction}
\begin{figure*}[!t]
\centering
\includegraphics[width=\linewidth]{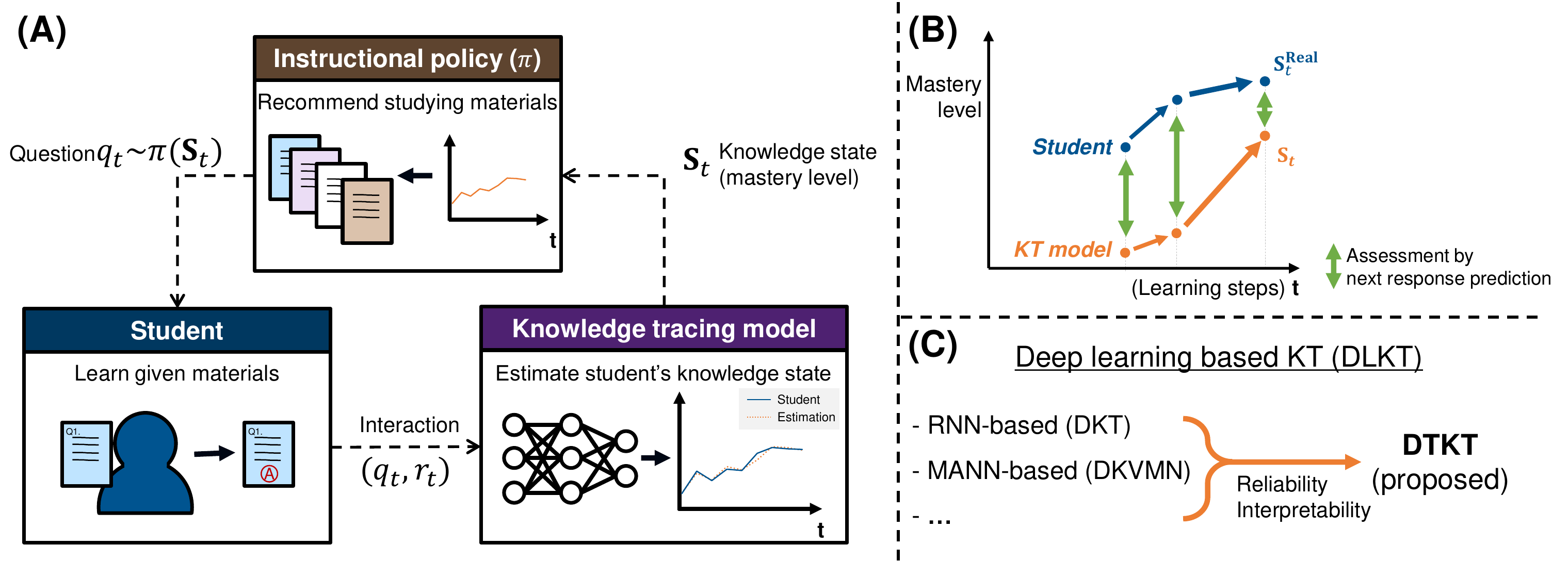}
\caption{
(A) Overview of the intelligent tutoring system (ITS). KT estimates the knowledge state of a student from the past interactions. An instructional policy recommends an appropriate question to the student based on the estimated knowledge state. The student's response to the given question (new interaction) is analyzed by KT. 
(B) The goal of KT is to estimate the mastery level of the student. 
(C) Our method (DTKT) improves the existing DLKT models in terms of reliability and interpretability.}
\label{fig:its_overview}
\end{figure*}

Intelligent tutoring systems (ITSs)~\cite{its_book,mooc_research} are interactive educational systems that provide personalized education to students. 
As shown in Fig.~\ref{fig:its_overview}(A), ITSs comprise three major components in general: a knowledge tracing (KT) model, an instructional policy model, and a student.
% Among them, KT model estimates the knowledge state of the involved student.

As shown in Fig.~\ref{fig:its_overview}(B), the goal of KT is to trace the knowledge state or mastery level of a student. % KT 설명, AUROC
% To build a successful ITS, it is essential to precisely estimate the student's mastery level. 
% Therefore, KT should provide accurate and trustworthy mastery level estimation. 
% $\mathbf{S}^{True}$ by the estimated knowledge state $\mathbf{S}$. 
Since it is difficult to directly estimate the real knowledge state of a student, KT needs a substitute method to evaluate the difference between real and estimated knowledge states.
Therefore, KT aims to minimize the loss for a binary classification that predicts the student's response (correct/incorrect) to the next question based on his/her past interactions.
In general, KT with high performance in terms of next response prediction is considered to provide better mastery level estimation.
% Thus, KT performs a binary classification which predicts the response (correct/incorrect) on a next question given the past performance to assess knowledge state indirectly. 
% KT with the higher area under the receiver operating curve (AUROC), which is the evaluation metric for the next response prediction, is considered to provide better mastery level estimation.

% 
%% 기존 KT의 문제점
There have been studies on deep learning-based KTs (DLKTs)~\cite{dkt,dkvmn}, and they have achieved higher predictive performance compared with traditional KT. 
The excellence of DLKT is attributed to the neural architecture, shared model, and interleaved input sequence~\cite{bkt_dkt_skill}.
However, there has been few studies that focused on the side effects of these factors.
% which make high performance on next response prediction no longer guarantee good mastery estimation. 
In this work, we introduce two side effects of DLKT, namely, the knowledge state update failure and the catastrophic forgetting.
These side effects make high predictive performance no longer guarantee good mastery estimation. 
These problems undermine the reliability of DLKT and render DLKT incompatible for an ITS. 
Furthermore, we observed that the non-interpretability of DLKT, attributed to these side effects, prevents DLKT from being extended to other educational scenarios.
% 
% of unintended forgetting in which these factors contribute to a number of side effects.
%Deep learning-based KT (DLKT)~\cite{dkt, dkvmn}은 traditional KT에 비해 우수한 AUROC를 보이며 이에 대한 원인으로는 neural network architecture, combined model, combined sequence (순서?)~\cite{}로 제안됐다. 그러나 이러한 요인들이 unintended forgetting을 발생시켜 여러 side effect를 일으키는 것에 대한 연구는 아직 없다. 
% 
%게다가 이러한 문제점들이 DLKT의 reliability를 파괴하여 ITS에 incomptabile하고 다른 educational applcation으로의 확장성을 제한하는 것을 보인다. 

As shown in Fig.~\ref{fig:its_overview}(C), we propose a novel KT model called Deep Trustworthy Knowledge Tracing (DTKT) to address the limitations of existing approaches. 
We extend the KT formulation from a sequence of sub-tasks to a sequence of pseudo multi-tasks, and propose a conditional pseudo-labeled loss. 
We experimentally demonstrate that the proposed method effectively alleviates the side effects of DLKT. 
We also find that the proposed method improves DLKT's reliability and interpretability, thereby making it compatible with an ITS and other educational applications. 
% In short, the proposed method is trustworthy KT.
The contributions of this paper are summarized as follows: 
%IKT는 KT 문제를 a sequence of sub-task form에서 a sequence of pseudo multi-task form으로 확장하며 conditional pseudo-labeled loss를 제안한다. 우리는 실험을 통해 the proposed method가 side effect들을 효과적으로 제거함을 확인하였다. 이로 인해 TKT는 높은 신뢰성과 interpretability를 가지게 되며 ITS 및 다른 educational application에 compatible 해진다. 
\begin{enumerate}
%     \item 
% We not only compare the differences between traditional KT and DLKT, but also provide an insight into the unique properties of DLKT compared to other machine learning problems.
    \item
We identify side effects of DLKT such as knowledge state update failure and catastrophic forgetting. 
They make DLKT unreliable to be used for educational services, and they are difficult to capture using the performance metric for next response prediction.
	\item 
To address the limitations of DLKT, we extend KT formulation from a sequence of sub-tasks to a sequence of pseudo multi-tasks from the insight of characteristics of DLKT. We introduce a novel regularization technique called a \textit{conditional pseudo-labeled loss}.
    \item 
Extensive experiments show that the proposed method regularizes the side effects, and make DLKT trustworthy. 
Our method provides reliability in mastery estimation and interpretability of DLKT. 
% in terms of an educational context (learning/forgetting). 
We also show that the proposed model trained under a scenario with forgetting can be extended to a scenario without forgetting.
\end{enumerate}

\section{Related Work}
\textbf{Traditional KT}
For performing student modeling, there are many machine learning-based methods, such as regression-based methods~\cite{irt,pfa}, matrix factorization-based methods~\cite{fm_kt}, and Bayesian network-based methods~\cite{bkt,dbn_kt}. 
Bayesian knowledge tracing (BKT)~\cite{bkt}, which is a prominent traditional KT method, uses a Hidden Markov Model (HMM) for each knowledge concept (\eg addition, and multiplication). 
The drawback of BKT is that it cannot model the relationship between the concepts. 
Some studies attempted to learn the relationships between the concepts; however, they were limited because either there were only a few concepts, or the relationship between the concepts had to be explicitly labeled ~\cite{dbn_skill_interaction,skill_combination}.
% If a question set has N concepts (\eg addition, multiplication), BKT has N distinct instances of HMM, assuming that each question implies only one concept. 
% This property inherently prevents BKT from learning multiple concepts from a given question, resulting in limited representation of the students' learning process.

\begin{figure*}[!t]
\centering
\includegraphics[width=\linewidth]{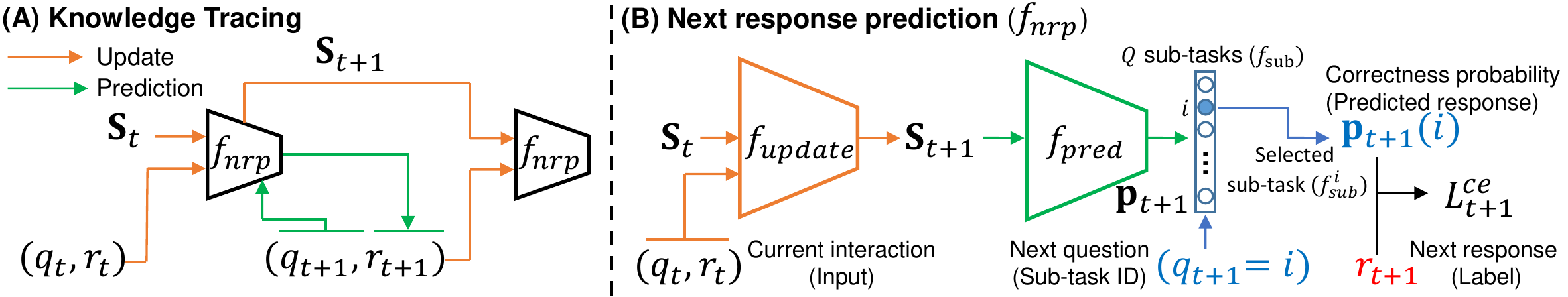}
\caption{
(A) Overview of knowledge tracing. $f_\textrm{nrp}$ is the next response prediction task. Orange and green lines mean $f_\textrm{update}$ and $f_\textrm{pred}$, respectively.
A current interaction $(q_t,r_t)$ is applied to the knowledge state $\mathbf{S}_{t+1}$ to predict the next response $r_{t+1}$. 
(B) $f_\textrm{update}$ updates $\mathbf{S}_t$ to $\mathbf{S}_{t+1}$ based on the current interaction $(q_t, r_t)$. $f_\textrm{pred}$ inferences the correctness probability vector $\mathbf{p}_{t+1}$.
There are $Q$ sub-tasks (\ie $f_\textrm{sub}$), but only one sub-task $f_\textrm{sub}^i$  selected by $q_{t+1}=i$ is performed at time step $t$.
The objective of KT model is to minimize cross entropy loss $L_{t+1}^{ce}$ between predicted response $\mathbf{p}_{t+1}(i)$ and $r_{t+1}$ given $q_{t+1}$. 
Note that $f_\textrm{nrp}$ and $\mathbf{S}_t$ can be implemented by various architectures such as HMM (BKT), LSTM (DKT), MANN (DKVMN).
}
\label{fig:kt_overview}
\end{figure*}

\textbf{DLKT} 
% 장점 및 richer information 
Unlike traditional KT, DLKT can learn the relationships between multiple concepts with higher predictive performance. 
% DLKT shows a higher predictive performance than that of traditional KT.
Deep knowledge tracing (DKT)~\cite{dkt} and its variants~\cite{kt_aaai_2018,dkt_hete,dkt_pre} are based on long short-term memory (LSTM)~\cite{lstm}, and dynamic key value memory network (DKVMN)~\cite{dkvmn} and its variants~\cite{hint_taking,deep_irt} are based on memory-augmented neural networks (MANNs)~\cite{mann,ntm}.
These DLKTs have enough model capacity, so that they can learn the relationships between various concepts. 
Furthermore, DLKT can utilize richer information such as the text of a question~\cite{kt_aaai_2018}, an image~\cite{similar_ex}, a hint~\cite{hint_taking}, and time information~\cite{dataforgetting}.

\textbf{Reliability}
Even if DLKT shows high performance on next response prediction, we found that DLKT does not provide reliable mastery level estimation. 
This prevents DLKT from being applied to ITS successfully.
Few works including the work in this paper, address the reliability of DLKT.
% In this paper, we aim to enhance the reliability of DLKT.

Our method is closely related to reconstructive regularization ~\cite{addressing}.
% However, the proposed method and reconstructive regularization are different.
However, our method shows promising reliability compared to reconstructive regularization.
This attributes to the properties of our method such as considering educational context, and effective regularization.
Furthermore, we provide novel insights into the side effects of DLKT from various perspectives. 
The detailed comparison between the proposed model and the reconstructive regularization is described in Supplementary Sec. 2. 

% KT는 학생의 지식 상태를 모델링하는 것이므로 실제 학생과 유사하다고 믿을 수 있어야 한다. 
% traditional KT는 모델이 상대적으로 간단하고 직관적이므로 researcher가 설계한대로 동작한다. 
% 반면 DLKT는 모델이 매우 복잡하여 내부적으로 일어나는 일에 대해 아직 깊은 이해가 부족한 상태이다. 
% 실제 DLKT를 보면 unintended behaviour를 보이는 경우가 존재하며 이러한 것은 DLKT의 reliability를 헤쳐 ITS에 사용할 수 없게 만든다. 
% DLKT의 unintended behavior에 초점을 맞추는 연구로는 ~\cite{addressing} 하나 존재한다. 
% Our work is closely related to the regularization term in ~\cite{addressing}.
% The authors of ~\cite{addressing} introduced \textit{reconstruction problem} which is similar in concept to \textit{update failure}, and proposed a regularization term \textit{r} as the solution. 
% 하지만 ~\cite{addressing}은 unintended behaviour를 완벽하게 해결하지 못하는 반면 우리는 거의 완벽하게 해결한다. 
% 더 나아가 우리는 이러한 limitation이 일어나는 것에 대해 다양한 측면에서 깊은 분석을 수행하였으며 DLKT의 interpretability와 applicability가 향상됨을 보였다. 
% Our proposed method와 ~\cite{adding}의 자세한 비교는 Supplementary A에 있다. 

% ~\cite{data_forgetting}이 forgetting을 다루지만 model level이 아니라 data 수준에서 다룬다. 
% The concepts of forgetting in the previous work~\cite{data_forgetting} and our work are different.
% The our method aims to reduce the unintended forgetting that occurs on neural architecture level. 
% However, ~\cite{data_forgetting} uses auxiliary time data as richer information and models forgetting based on time data. 

\textbf{Interpretability}
To understand the characteristics of KT, the behavior of each component of the KT model should correspond to the behavior of real student (\eg learning, forgetting, slipping, and guessing). 
Several studies concentrated on the interpretability of traditional KT ~\cite{time_matter,slip_guess_bkt,irt}.
% \textcolor{red}{Although DKT has better predictive performance, it is not interpretable from the educational context~\cite{dkt_interpretability,how_deep}.}
Some studies addressed the interpretability of DLKT in terms of the input data~\cite{dkt_hete}, difficulty level of the question~\cite{deep_irt}, and concept of the question~\cite{dkvmn}.
However, no studies, so far, exist on the interpretability of DLKT from the perspectives of learning and forgetting, and this paper is the first toward that direction.

\citeauthor{dataforgetting} address the forgetting pattern of the student by using auxiliary time information.
This method is similar to our method in terms of considering the forgetting pattern in DLKT.
% ; however the previous work 
Our method regularizes the unintended forgetting of the components to make DLKT trustworthy. 
In contrast, the previous work just aim to learn the forgetting of the student based on the additional time information, and does not take account of interpretability of its components.  
% In this paper, we do not use additional information except (question, response) pair.

% Suppl에 표로 만들어서 넣을까?  (바로 만들어서 넣자. neurips 버전에도 넣을 수 있을 것) 
% KT의 complexity를 너무 많이 제한해버려서 실제 ITS에서 일어날 수 있는 시나리오 또한 제한함. 
% 솔루션이 간접적인 해결책, 문제 자체를 해결하지 못해서 솔루션을 적용해도 문제가 남아 있음. 
% 통계적 기법으로만 분석, probolem이 일어나는 이유를 정성적 실험으로 보임 
% dataset의 imbalance 성질을 고려함. 
% DLKT 문제를 일반화 및 다른 ML application들과의 비교 
% Interpretability 측면으로 접근함. 
% 학생의 forgetting을 고려함. 
% Catastrophic forgetting 측면에서 분석
% 새로운 시나리오를 가능하게 함 (learning with forgetting --> testing with non-forgetting) 

% waiveness가 비슷한 구조를 가지고 있음.
% 아래와 같이 설명하고 있으나 교육학적으로는 의미가 없음? 우리는 몇 가지 가능한 시나리오를 허용하고 있음. 
% 이렇게 무조건적으로 제한하는 건 문제 간 유의미한 interaction 모델링을 포기하는 것. 
% This is not desirable because student’s knowledge state is expected to transit only gradually and steadily over time

% Both methods aim to regularize the $\mathbf{p}_{t+1}(i)$ given $(q_t=i, r_t)$. 
% However, given correct answer $(q_t=i,r_t=1)$, [P1] gives a penalty to the model when $\mathbf{p}_{t+1}(i) \ne 1$ while our term gives a penalty when $\mathbf{p}_{t+1}(i)-\mathbf{p}_{t}(i) < 0$.
% The regularization in [P1] does not explicitly prevent a decrease in probability; thus, in that case, it is possible that knowledge state update failure exists. As shown in Table A, [P1] shows non-zero mastery update failure ratios, meaning that mastery failures still remain.

\section{KT Formulation} \label{sec:formulation}

%% 환경 설명 
An instructional policy $\pi$ provides a question ID $q_t \in [1,Q] \sim \pi$ to a student, where $Q$ denotes the number of all the questions.
The student gives a response $r_t \in \{0,1\}$, which has been sampled from the mastery distribution $p(r_t|\mathbf{S}_t^{\mathrm{Real}},q_t)$ where $\mathbf{S}_t^{\mathrm{Real}}$ denotes the real knowledge state of the student.
% The student gives a response $r_t \in \{0,1\}$, which is sampled from the mastery distribution $p(r_t|\mathbf{S}_t^{True},q_t)$ (\ie $r_t \sim p(r_t|\mathbf{S}_t^{True},q_t)$) where $\mathbf{S}_t^{True}$ is the student's real knowledge state.
% The student gives a response $r_t \in \{0,1\}$, which is sampled from the mastery distribution $p(r_t|\mathbf{S}_t^{True},q_t)$ (\ie $r_t \sim p(r_t|\mathbf{S}_t^{True},q_t)$) where $\mathbf{S}_t^{True}$ is the student's real knowledge state.

%% KT의 목적 
We define an interaction $x_t = (q_t,r_t)$ and a sequence of interactions $x_{1:T}$, where $T$ denotes the terminal time step. 
The ultimate goal of KT is to estimate $\mathbf{S}_t^{\mathrm{Real}}$ and $p(r_t|\mathbf{S}_t^{\mathrm{Real}},q_t)$ the given $x_{1:T}$.

%% 실제 KT가 어떻게 구현되는지.
At time step $t$, KT performs a binary classification task to predict the response $r_{t+1}$ to the new question $q_{t+1}$ given past interactions $x_{1:t}$~\cite{bkt,dkt,dkvmn}. 
To solve this task, KT utilizes a knowledge state estimator $\mathbf{S}_t$ for tracing $\mathbf{S}_t^\mathrm{Real}$. 
We define the next response prediction task $f_\textrm{nrp}$ as follows :
\begin{align}
    \label{eq:nrp}
    (\mathbf{p}_{t+1}, \mathbf{S}_{t+1}) \leftarrow f_\textrm{nrp}(\mathbf{S}_{t}, (q_{t}, r_{t})),
\end{align}
where $\mathbf{p}_{t+1} \in \mathbb{R}^{Q}$ denotes the correctness probability vector, whose element $\mathbf{p}_{t+1}(i) \in \mathbb{R}$ denotes $p(r_{t+1}=1|\mathbf{S}_{t+1},q_{t+1}=i)$.
Then, KT can be viewed as a sequence of $f_\textrm{nrp}$ as depicted in Fig.~\ref{fig:kt_overview}(A). Fig.~\ref{fig:kt_overview}(B) represents the two components of $f_\textrm{nrp}$, namely, the update function $f_\textrm{update}$ and the prediction function $f_\textrm{pred}$, both of which are defined as follows:
% \begin{align}
%     &\mathbf{S}_{t+1} \leftarrow f_\textrm{update}(\mathbf{S}_{t}, (q_{t}, r_{t})), &\\
%     &\mathbf{p}_{t+1} = f_\textrm{pred}(\mathbf{S}_{t+1}).&
% \end{align}
\begin{align}
    &\mathbf{S}_{t+1} \leftarrow f_\textrm{update}(\mathbf{S}_{t}, (q_{t}, r_{t})), \\
    &\mathbf{p}_{t+1} = f_\textrm{pred}(\mathbf{S}_{t+1}).
\end{align}
% Given current interaction $(q_t, r_t)$, $f_\textrm{update}$ updates the $\mathbf{S}_t$ to $\mathbf{S}_{t+1}$.
% If $r_t=1$, KT updates $\mathbf{S}_{t}$ in the direction of increasing in mastery level of $q_t$. 
% Given $\mathbf{S}_{t+1}$, $f_\textrm{pred}$ calculates $\mathbf{p}_{t+1}$.

% \textbf{Inference} 
% % Given $\mathbf{S}_{t+1}$, KT calculates $p_{t+1} \in \mathbb{R}$ which means the probability for answering $q_{t+1}$ correctly as follows:
% \begin{align}
%     % \mathbf{p}_{t+1} = f_\textrm{pred}(\mathbf{S}_{t+1}) = f_\textrm{pred}(f_\textrm{update}(\mathbf{S}_t, (q_t,r_t))),
% \end{align}
% where $f_\textrm{pred}(\cdot)$ denotes the prediction function. 

\textbf{Training}
The parameters of $f_\textrm{update}$ and $f_\textrm{pred}$, and the initial knowledge state $\mathbf{S}_0$ are trained to minimize the cross-entropy loss. Given a next question $q_{t+1}=i$ and student's real response $r_{t+1}$, the cross entropy loss between the prediction $\mathbf{p}_{t+1}(i)$ and the label $r_{t+1}$ is calculated as follows:
\begin{align}
    L_{t+1}^{ce} = -r_{t+1} \log (\mathbf{p}_{t+1}(i)) - (1-r_{t+1}) \log (1-\mathbf{p}_{t+1}(i)).
    % L^{ce} = \sum_{t=0}^T L_{t+1}^{ce} = \sum_{t=0}^T [ -r_{t+1} \log (\mathbf{p}_{t+1}(i)) - (1-r_{t+1}) \log (1-\mathbf{p}_{t+1}(i)) ]. 
    % L_{t+1}^{ce} = -r_{t+1} \log (\mathbf{p}_{t+1}(i)) - (1-r_{t+1}) \log (1-\mathbf{p}_{t+1}(i)), 
    \label{eq:ce_loss}
\end{align}
% where $p_{t+1} = \mathbf{p}_{t+1}(i)
% The performance of KT is evaluated by AUROC. 
Note that $r_t$ has different roles for each component of the KT model. $r_t$ denotes the direction of the input $(q_t,r_t)$ in $f_\textrm{update}$, and label of binary classification in $f_\textrm{pred}$.

% Since the ultimate goal of KT is to estimate $\mathbf{S}_t^{True}$, the ideal KT is supposed to predict response of \textbf{any} next question.
At time step $t$, the ideal goal of KT is to predict the response to \textit{any} next question, then $f_\textrm{nrp}$ can be considered as a multi-task function for the total number $Q$ of the questions. 
However, because of the sequential form of the input, KT can acquire only one label $r_{t}$ on $q_t=i$ at time step $t$.
KT can be viewed as a sequence of sub-tasks upon defining sub-task $f_\textrm{sub}^i$ as follows:
\begin{align}
    f_\textrm{sub}^i(\mathbf{S}_{t}, (q_{t}, r_{t})) = f_\textrm{nrp}(\mathbf{S}_{t}, (q_{t}, r_{t})|q_{t+1}=i),~ \forall i \in [1,Q],
\end{align}
where $ (\mathbf{p}_{t+1}(i), \mathbf{S}_{t+1}) \leftarrow f_\textrm{sub}^i(\mathbf{S}_{t}, (q_{t}, r_{t}))$.
The ID of the next question $q_{t+1}$ is equal to the ID of the sub-task.
% From this perspective, a next question ID $q_{t+1}$ is equal to the sub-task ID.

\section{Deep Trustworthy Knowledge Tracing}
In this section, we introduce the side effects of the existing DLKT and propose a Deep Trustworthy Knowledge Tracing (DTKT) to address these side effects.

\begin{figure*}[!t]
\centering
\includegraphics[width=\linewidth]{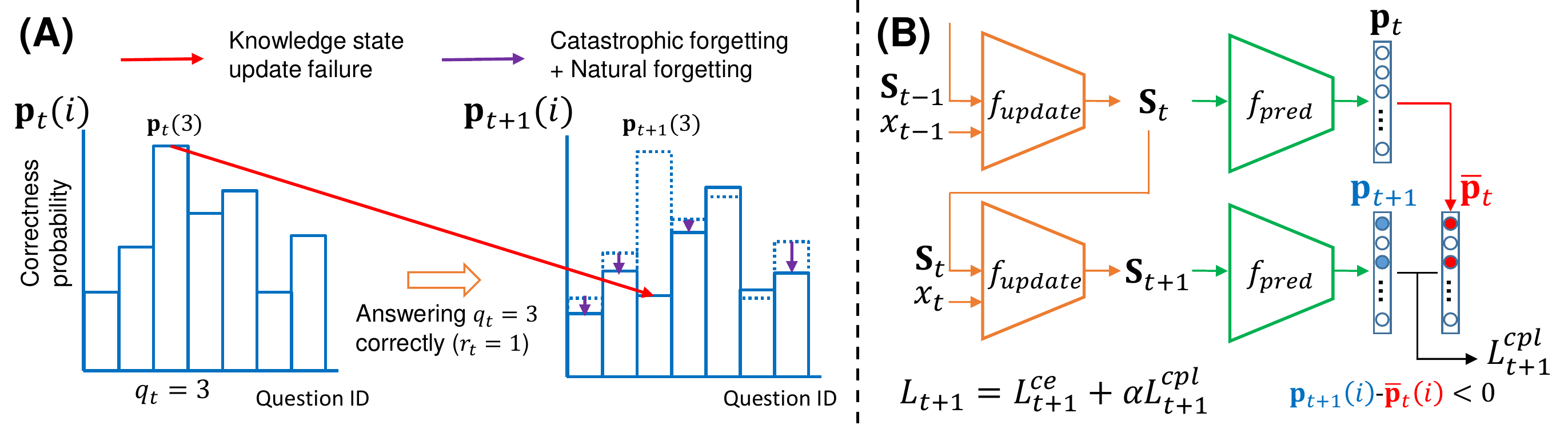}
\caption{(A) Overview of side effects: knowledge state update failure (red) and catastrophic forgetting (purple). For example, before answering $q_t=3$ correctly, the correctness probability $\mathbf{p}_t(3) = 0.9$. After answering $q_t=3$ correctly, the correctness probability $\mathbf{p}_{t+1}(3) = 0.3446$. Therefore, $0.5444~(|0.9 - 0.3446|)$ is the maximum observed probability decrease in this case. Probability decrease for other questions are caused by catastrophic forgetting and natural forgetting. (B) Overview of our method (DTKT).}
\label{fig:side_effects}
\end{figure*}

\subsection{Side effects of the existing DLKT}
Unlike conventional KT, which implements $f_\textrm{nrp}$ and $\mathbf{S}_t$ for each concept by using a separate model, DLKT implements $f_\textrm{nrp}$ and $\mathbf{S}_t$ for all the concepts by using a single LSTM~\cite{dkt,dkt_hete} or a single MANN~\cite{dkvmn,hint_taking}. 
These models address the total sub-tasks jointly (shared model), and receive an input consisting of various knowledge concepts (interleaved sequence). 
Based on a comparative study between DLKT and traditional KT~\cite{bkt_dkt_skill}, the higher predictive performance of DLKT can be attributed to the shared model and interleaved sequences.
However, these approaches can lead to side effects such as the knowledge state update failure and the catastrophic forgetting.
These side-effects reduce the reliability of KT in mastery level estimation even if KT shows high performance on next response prediction.
% These side-effects are difficult to capture by AUROC, and reduce the reliability and interpretability of DLKT.

LSTM and MANN showed high performance in modeling the sequential data~\cite{lstm,mann}, and we assume that their capacity is large enough to model the natural forgetting of a student.
However, we observed forgetting patterns which do not correspond to real human forgetting, and we call these patterns as unintended forgetting.
DLKT learns the unintended forgetting as well as the natural forgetting, and this prevents DLKT from being extended to various educational applications.

\textbf{Knowledge state update failure}
We define the \textit{knowledge state update failure} (or, simply, \textit{update failure}) as information of an interaction $(q_t=i,r_t=1$) is applied improperly when updating $\mathbf{S}_t$ to $\mathbf{S}_{t+1}$. 
We can observe the update failure when estimated mastery level of the question decreases even if a student answers that question correctly (\ie $\mathbf{p}_{t+1}(i)-\mathbf{p}_{t}(i) < 0$ given $(q_t=i,r_t=1)$, see Fig.~\ref{fig:side_effects}(A)).
KT with update failure exploits information from $(q_t=i,r_t=1$) in a wrong direction, and cannot provide reliable mastery estimation. 

The power of DLKT is derived from the ability to comprehend the relationships among several concepts, and a knowledge state $\mathbf{S}_t$ stores the mastery level of all concepts.
DLKT assumes that a question has multiple concepts, and it trains the concept distribution of each question automatically.
% Fig. ~\ref{fig:side_effects} A) represents knowledge state update with concept distribution. 
Information of $(q_t=i,r_t=1)$ should be used to update the part of $\mathbf{S}_t$ related to the concept distribution of $q_t$.
However, given an interleaved sequence with multiple concepts, knowledge state update failure is caused by a mismatch between concepts of current $q_t$ and next $q_{t+1}$.
$L^{ce}_{t+1}$ aims that $(q_t=i,r_t=1)$ is used for $\mathbf{S}_t$ related to $q_{t+1}$.
If the concept distribution of $q_t$ and $q_{t+1}$ does not share common concepts (mismatched), information of $(q_t=i,r_t=1)$ is not properly updated.
We did proof-of-concept experiment in Section 5.1 to prove our claim.

Even if DLKT has the update failure problem, predictive performance can be high because it is evaluated by the response prediction for the next question, not current question. 
As a result, high performance on next response prediction does not imply good mastery estimation in DLKT, and it should be verified that DLKT has no update failure.

\textbf{Catastrophic forgetting} 
% \textcolor{red}{Continual learning 쪽 catastrophic forgetting 이야기 추가}
To calculate $L^{ce}_{t+1}$, only one sub-task is selected by $q_{t+1}$, because KT is a sequence of sub-tasks. 
We define \textit{catastrophic forgetting} as the information loss of other sub-tasks by training a selected sub-task.
% Since total $Q$ sub-tasks share parameters in DLKT, training one sub-task affect other sub-tasks. 
Since all sub-tasks in DLKT share a single model, training a given sub-task can disturb trained parameters for other sub-tasks, as shown in Fig.~\ref{fig:side_effects}(B).
Disturbed parameters hinder the knowledge state from being updated to the right direction.
Similar to the update failure, DLKT with catastrophic forgetting can still have high performance on next response prediction. 

\textbf{Non-interpretability}
Both LSTM-based and MANN-based KT models have forgetting components (\eg forget gate and erase operation) and learning components (\eg input gate and add operation). 
However, learning and forgetting components of existing DLKT does not correspond to natural learning and forgetting of a real student, respectively, since DLKT learns not only natural forgetting but also unintended forgetting due to update failure and catastrophic forgetting. 
If there is a right correspondence between DLKT and real student, which means DLKT is interpretable in terms of educational context, DLKT can be applied to other scenarios which are different from the scenario that the model was trained for. 
For example, the explainable DLKT model trained for a scenario with forgetting (\eg one semester lecture) can be easily extended to the scenario without forgetting (\eg one hour test) by disabling the forgetting component of the DLKT.
If there are small number of data collected from the non-forgetting scenario while there are large number of data gathered from the forgetting scenario, this property may be very useful. 
\ctable[
star,
pos = t,
center,
caption = {The number of students, questions, and interactions in the three datasets used in the experiments. Count means how many times a question is addressed in the entire interaction.},
%mincapwidth = \textwidth,
%width=\textwidth,
label = {tab:data},
doinside = {\footnotesize \def\arraystretch{.7}}
%]{lccccccc[table-format=2.3]}{
]{lrrrrrr}{
%     \tnote[a]{ calculated by using data from~\cite{horowitz20141}}
}{
\toprule
Datasets & \# of students & \# of questions & \# of interactions & max count & min count  \\
\midrule
Synthetic-5     &            4,000  &                 50  &       200,000 & 4,000    & 4,000   \\
ASSISTments2009 &            4,151  &                110  &       325,637 & 24,253   & 5       \\
% ASSISTments2015 &           19,840  &                100  &       683,801 & 21,694   & 265     \\
Statics2011     &              333  &              1,223  &       189,297 & 304      & 2       \\
\bottomrule
}

\subsection{Conditional pseudo-labeled loss}
Update failure occurs when information of current interaction is applied to knowledge state in a wrong direction due to the mismatched concepts between current (input) and next question (label).
To prevent this update, a pseudo label which has the same concept as current question can be used.

To estimate the mastery level correctly, KT should predict the response of \textit{any} question, but KT focuses on only the given question.
Catastrophic forgetting occurs when training processes of the given sub-task disturb the trained parameters for the other sub-tasks. 
To address this problem, other sub-tasks can have a pseudo label that preserves their information.  
% Additionally, in the given sub-task, mismatched concepts between current and next questions cause the update failure. We found the problem is that the current question is not considered in current update. (----)
% and the next question should have similar concepts to the previous question to address this limitation. 

Therefore, we extend the KT formulation from a sequence of sub-tasks to a sequence of pseudo multi-tasks. We propose the conditional pseudo-labeled loss $L^{cpl}$ as follows:
\begin{equation}
\label{eq:cpl_loss}
\begin{split}
L_{t+1}^{cpl} = \sum_{j}^Q & |\bar{\mathbf{p}}_{t}(j) - \mathbf{p}_{t+1}(j)|^2 \\
& \cdot \mathbbm{1} (\mathbf{p}_{t+1}(j) - \bar{\mathbf{p}}_{t}(j) <0  |   q_{t}=i, r_{t}=1)
\end{split}
\end{equation}
where $i,j \in [1, Q]$ are the question IDs, $\bar{\mathbf{p}}_{t}(i)$ is a pseudo-label at time step $t$, and $\mathbbm{1} \in [0,1]$ denotes an indicator function.
Note that $\mathbf{p}_{t+1} = f_\textrm{pred}(\mathbf{S}_{t+1})$, whereas $\bar{\mathbf{p}}_{t} = f_\textrm{pred}(\mathbf{S}_{t})$.
Then, the parameters of DTKT are trained to minimize  $L_{t+1}^{ce} + \alpha L_{t+1}^{cpl}$, where $\alpha$ is a hyperparameter.

For each question, mismatch of concepts is resolved by using the pseudo label $\bar{\mathbf{p}}_{t}(j)$ for $\mathbf{p}_{t+1}(j)$. 
Catastrophic forgetting is solved by providing pseudo labels for questions suffering from the unintended forgetting.

Intuitively, $L_{t+1}^{cpl}$ makes each sub-task preserve its own previous information. 
This means that $L_{t+1}^{cpl}$ can prevent interactions among sub-tasks. 
To allow only positive interactions among sub-tasks, pseudo-label $\bar{\mathbf{p}}_{t}(j)$ is applied only when $(\mathbf{p}_{t+1}(j) - \bar{\mathbf{p}}_{t}(j) <0 \ | \ q_{t}=i, r_{t}=1)$.
It is possible to utilize various conditions, and we performed an ablation study for several conditions (Supplementary Sec. 3).

\section{Experimental Results}\label{result}
We used three public datasets: simulated data (Synthetic-5) from a previous work \cite{dkt}, skill-builder datasets of ASSISTments~\cite{data_assist}, and the OLI Engineering Statics dataset\footnote{https://pslcdatashop.web.cmu.edu/DatasetInfo?datasetId=507}. 
The statistics of each dataset are presented in Table~\ref{tab:data}. 
Instead of using richer information, DTKT uses only $(q,r)$ pairs.  

We selected a DKVMN~\cite{dkvmn} as the base model of DLKT, and its architecture is described in Supplementary Sec. 4.
DKVMN is a state-of-the-art KT model in terms of predictive performance and it is also interpretable in terms of the concept of a question.
We performed a proof-of-concept experiment by using this property of the DKVMN (the concept-interpretability).
We evaluated its predictive performance using area under the receiver operating characteristic curve (AUROC), and the detailed results are presented in Supplementary Sec. 1.
DTKT with suitable $\alpha$ shows almost the same performance on AUROC as the base DLKT. 
The performance gaps between the base and proposed models are less than 0.005. 
This suggests that the proposed method increases the reliability of DLKT without degradation of predictive performance.

To measure the effect of both the update failure and the catastrophic forgetting, we define a probability difference matrix $\Delta \mathbf{p}_t (\cdot | \cdot) \in \mathbb{R}^{Q\times Q}$ and the average decrease in the mastery level as $md_t \in \mathbb{R}$ as follows:
% inan interaction $(q_t=i, r_t=1)$ as follows: 

\begin{equation}
\label{eq:pd}
\begin{split}
\Delta \mathbf{p}_t(j|i) = ~&\mathbf{p}_{t+1}(j) - \mathbf{p}_{t}(j) \\
&\textrm{given} ~~ (q_t=i,r_t=1), \forall i,j \in [1,Q], 
\end{split}
\end{equation}

% \begin{equation}
% \label{eq:md}
% \begin{split}
% md_t = \frac{1}{Q} \sum_i^Q &\frac{1}{\sum_j^Q \mathbbm{1} (\Delta \mathbf{p}_t(j|i) <0 )}|\Delta \mathbf{p}_t(j|i)| \\ 
% &\cdot \mathbbm{1} (\Delta \mathbf{p}_t(j|i) <0 ).
% \end{split}
% \end{equation}

\begin{equation}
\label{eq:md}
md_t = \frac{1}{Q} \sum_i^Q \frac{1}{\sum_j^Q \mathbbm{1} (\Delta \mathbf{p}_t(j|i) <0 )}|\Delta \mathbf{p}_t(j|i)| \cdot \mathbbm{1} (\Delta \mathbf{p}_t(j|i) <0 ).
\end{equation}

\subsection{Knowledge state update analysis}
We can observe the update failure where $\Delta \mathbf{p}_t(i|i) < 0$, and its effect can be evaluated by $|\Delta \mathbf{p}_t(i|i)|$.
In our experiment, we ignored the subtle probability difference which can originate from the randomness of neural networks (\ie we regarded only the condition $|\Delta \mathbf{p}_t(i|i)| > th$  where $th$ denotes the threshold).
Table~\ref{tab:ni_itself} presents the analysis on the update failure. 
We calculated the ratio of the questions having update failure to the total number of questions.
We also reported the conditional average and maximum value of $|\Delta \mathbf{p}_t(i|i)|$ when $|\Delta \mathbf{p}_t(i|i)| > th$. 
The maximum $|\Delta \mathbf{p}_t(i|i)|$ on the ASSISTment2009 dataset was observed to be $0.5554$, meaining that the mastery level became smaller than 0.5 even though the student answered correctly. 
On Statics2011, 28.62\% of the questions suffered from update failure.  

Applying $L^{cpl}$, we observed that model had no update failure on the Synthetic-5 and ASSISTment2009.
For Statics2011, the ratio decreases to 0.41\% when $\alpha=0.001$. 
As shown in Supplementary Sec. 1, the predictive performance does not decrease for all the models. 
$L^{cpl}$ effectively regularizes the wrongly directed knowledge state update without predictive performance degradation.

\ctable[
star,
pos = t,
center,
caption = {Analysis on the knowledge state update failure: $\Delta \mathbf{p}_t(i|i) < 0$. 
Ratio is the number of questions with the knowledge state update failure over the total number of questions. 
The average and maximum values of $|\Delta \mathbf{p}_t(i|i)|$ are reported.  
We filtered out $\Delta \mathbf{p}_t(i|i)$ when $|\Delta \mathbf{p}_t(i|i)| < th=0.001$ to ignore subtle differences. 
}, 
label = {tab:ni_itself},
doinside = {\footnotesize \def\arraystretch{.7}}
%]{lccccccc[table-format=2.3]}{
]{lrrrrrrrrr}{
}{
\toprule
Dataset & \multicolumn{3}{c}{Synthetic-5} & \multicolumn{3}{c}{ASSISTment2009} & \multicolumn{3}{c}{Statics2011} \\
\midrule
$\alpha$ & \multicolumn{1}{c}{ratio(\%)} & \multicolumn{1}{c}{average} & \multicolumn{1}{c}{max} & \multicolumn{1}{c}{ratio(\%)} & \multicolumn{1}{c}{average} & \multicolumn{1}{c}{max} & \multicolumn{1}{c}{ratio(\%)} & \multicolumn{1}{c}{average} & \multicolumn{1}{c}{max} \\
\midrule
0 (base) & 14.0 & 0.0869 & 0.2117 & 7.27 & 0.1633 & 0.5554 & 28.62 & 0.0088 & 0.0475 \\
0.0001 & 0  & 0  & 0  & 0  & 0  & 0  & 24.12 & 0.0037 & 0.0225 \\
0.001  & 0  & 0  & 0  & 0  & 0  & 0  & 0.41 & 0.0016 & 0.0026 \\
\bottomrule
}
\begin{figure}[t]
\centering
\includegraphics[width=1\linewidth]{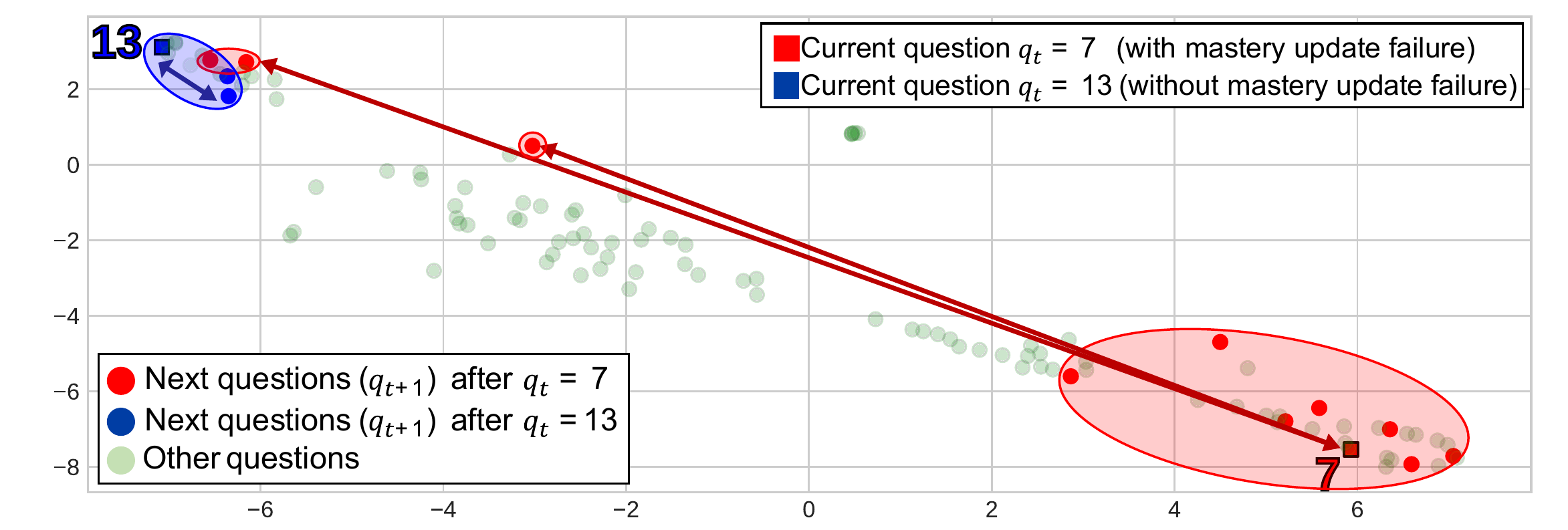}
\caption{Concept distribution of questions plotted by t-SNE. Consecutive questions with update failure (red points) have different concepts while those without update failure (blue points) have similar concepts.}
\label{fig:concept_dist}
\end{figure}

\textbf{Proof-of-concept for the cause of update failure}
We claim that the update failure can be attributed to the mismatched concept distributions both of $q_t$ and $q_{t+1}$.
We observed that $q_t=7$ suffered from update failure and $q_t=13$ did not, on the ASSISTment2009. 
We then transformed each question to the concept vector by using the component of DKVMN ($\mathbf{w}_{t+1}$, see Fig.B in Supplementary).  
We then plotted the concept vectors of the question by using t-SNE. 
As depicted in Fig.~\ref{fig:concept_dist}, some $q_{t+1}$ after $q_t=7$ have different concepts from that of $q_t=7$ (red), while the $q_{t+1}$ after $q_t=13$ have similar concepts with that of $q_t=13$ (blue).
We filtered out $q_{t+1}$ that appeared less than 15 times after $q_t$ in the training dataset (green).
This result shows that the update failure is caused by the mismatched concept of $q_t$ and $q_{t+1}$.

\begin{figure*}[!t]
\centering
\includegraphics[width=0.7\linewidth, height=4cm]{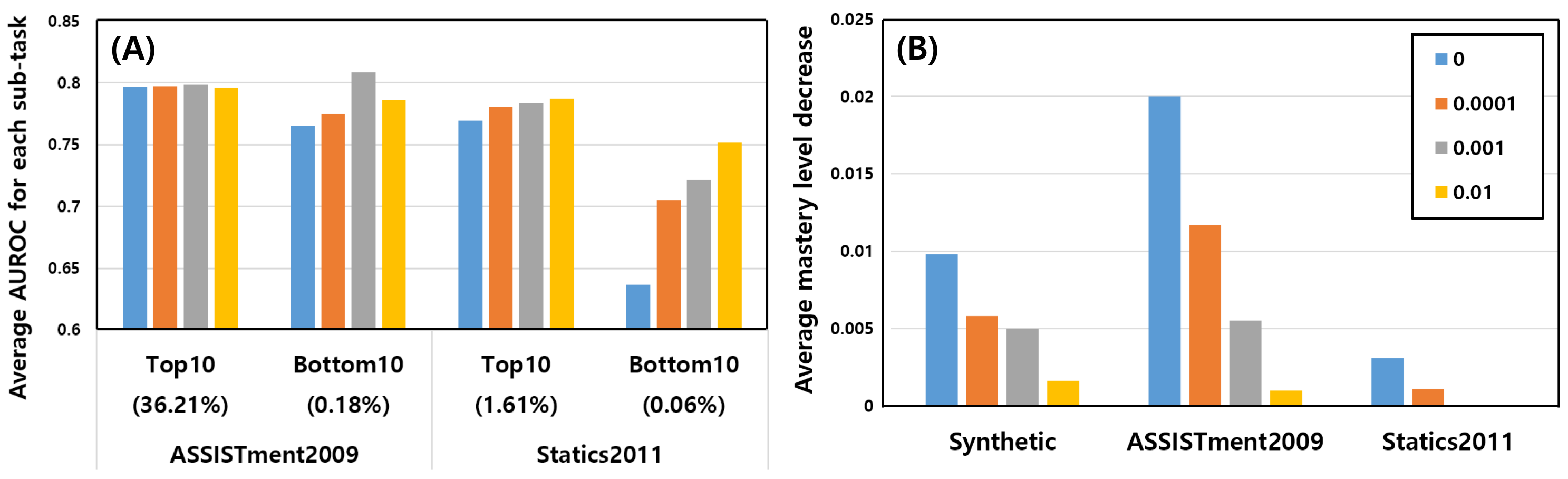}
\caption{Catastrophic forgetting analysis from the two perspectives. (A) Average AUROC for the sub-tasks corresponding to top 10 and bottom 10 questions in terms of count. The percentage means the ratio of the top 10 and bottom 10 to all the questions. % TODO check
(B) Each bar represents $md_t$ (see Eq.~\ref{eq:md}).}
\label{fig:catastrophic_forgettting}
\end{figure*}

\begin{figure*}[!t]
\centering
\includegraphics[width=0.9\linewidth, height=4cm]{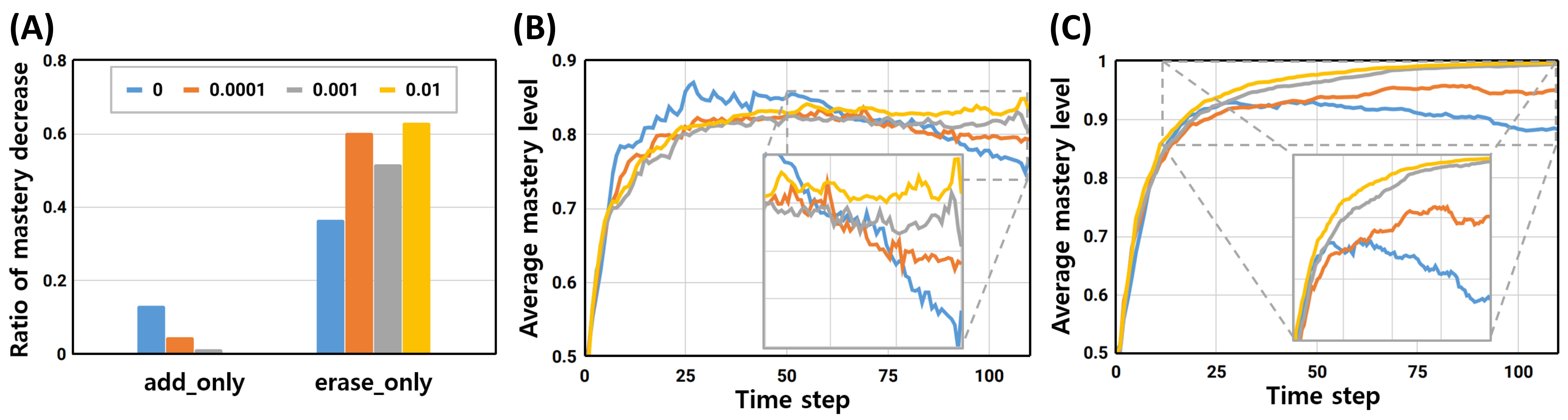}
\caption{Interpretability analysis on the ASSISTment2009 dataset. (A) The ratio of the number of $q_t=i$ to the total $Q$ questions when ($\Delta \mathbf{p}_t(j|i)<0, \, \forall \, j \neq i$). (B) and (C) plot ($\frac{1}{Q}\sum_{i=1}^Q \mathbf{p}_t(i)$) for $add+erase$ and $add\_only$, respectively.}
\label{fig:interpretability}
\end{figure*}

\subsection{Catastrophic forgetting analysis}\label{sec:imbalance}
We analyze the limitations caused by the catastrophic forgetting from the perspectives of AUROC for each sub-task, and correctness probability. 
Similar to the update failure, DLKT with high AUROC for all the questions can suffer from the limitation of catastrophic forgetting.

\textbf{Imbalanced AUROC for each sub-task}
We define the term \textit{count} as the number of times a question appears in the dataset. 
In Table~\ref{tab:data}, the maximum and minimum counts are presented, and there is a large difference between them, except for the Synthetic-5 dataset.
Fig.~\ref{fig:catastrophic_forgettting}(A) depicts the average AUROC for the sub-tasks corresponding to the top 10 and bottom 10 questions in terms of count.
The blue bar represents the base model, and the others are the proposed models having various values of $\alpha$. 

For the ASSISTment2009 dataset, the top 10 questions account for 36.22\% of the dataset. Since the sub-tasks corresponding to the top 10 questions have enough opportunities to be trained, all these sub-tasks have a similar AUROC. % TODO check
However, the sub-tasks for the bottom 10 questions (comprising only 0.18\% of the dataset) have few opportunities to be trained; even if these sparse sub-tasks train the parameters properly, the information of these sparse sub-tasks can be lost while frequent sub-tasks are being trained, because all the sub-tasks share parameters in DLKT. 
The base model has large difference of AUROC between the sub-tasks corresponding to the top 10 questions (0.797) and the bottom 10 questions (0.765), because of the catastrophic forgetting.
Applying the conditional pseudo labeled loss, the AUROC difference becomes smaller as $\alpha$ becomes larger, except for $\alpha=0.01$.
In the case of $\alpha = 0.001$, the average AUROC for the sub-tasks corresponding to bottom 10 questions (0.808) becomes comparable to that of the top 10 questions (0.798). 

For the Statics2011 dataset, the top 10 and bottom 10 questions in terms of counts account for 1.61\% and 0.06\% of the dataset, respectively. % TODO check
In the case of the base model, the average AUROCs for the sub-tasks corresponding to the top 10 and bottom 10 questions are 0.769 and 0.637, respectively. 
Applying the proposed method, the AUROC difference becomes smaller as $\alpha$ increases. 
The AUROC for the sub-task corresponding to the top 10 questions also increases because these top 10 questions comprise 1.61\% of the dataset, small enough to be affected by catastrophic forgetting. % TODO check

The results show that catastrophic forgetting destroys the information of the sub-tasks, and that the proposed method preserves the information of the sparse sub-tasks.  %preserve their parameters.
The sparse sub-tasks have little influence on the AUROC for total sub-tasks because of their sparseness. 
% Although the AUROCs of these sparse sub-tasks are smaller than those of frequent sub-tasks, 
Thus, DLKT models have similar AUROCs for all the sub-tasks regardless of low AUROCs of sparse sub-tasks. 

\textbf{Probability difference}
Fig.~\ref{fig:catastrophic_forgettting}(B) plots $md_t$ (see Eq.~\ref{eq:md}) which denotes the average value of mastery level decrease caused by both the natural forgetting of a student and the catastrophic forgetting of the model. 
As $\alpha$ becomes larger, $md_t$ becomes smaller. 
KT will have performance degradation in the AUROC if natural forgetting of KT is removed.
As shown in Supplementary Sec. 1, all models on the same dataset have comparable AUROCs, on the basis of which, it is demonstrated that the proposed method regularizes the catastrophic forgetting based negative influences among the sub-tasks but it does so while preserving natural forgetting. 

\subsection{Interpretability}
% \subsection{Interpretability from the perspective of educational context}
We tested the correspondence of the learning/forgetting components of DLKT to those of a real human. The case in which the forgetting component is disabled is referred to as $add\_only$, and the case in which the learning component is disabled is referred to as $erase\_only$ (see Eqs. 10 and 11 in the Supplementary).

Fig.~\ref{fig:interpretability}(A) depicts the percentage of the number of $q_t=i$ to the total $Q$ questions when $\Delta \mathbf{p}_t(j|i)<0, \, \forall \, j \neq i$.
For the base model ($\alpha=0.0$), the $add\_only$  decreases the correctness probability of other tasks even if the student answers $q_t=i$ correctly, meaning that the roles of the learning and forgetting components are not separated.
On $\alpha$ becoming larger, $md_t$ becomes larger for the $erase\_only$ , and $md_t$ becomes smaller for the $add\_only$. 
These results suggest that the proposed method trains the learning and forgetting components to correspond to real human learning and forgetting by regularizing both the update failure and the catastrophic forgetting. 

\subsection{Applicability to the other scenario}
The datasets used in this paper were collected under scenarios with forgetting. 
We simulated the trained model with forgetting ($add+erase$) can also be used to trace the mastery of a student for the non-forgetting scenarios ($add\_only$). 
For simple simulation, we assume that the questions are given in the descending order of difficulty, and that the student answers all the questions correctly.
The results in Figs.~\ref{fig:interpretability}(B) and (C) represent the average mastery level ($\frac{1}{Q}\sum_{i=1}^Q \mathbf{p}_t(i)$), and Fig.~\ref{fig:interpretability}(B) represents the scenario with forgetting (\eg lecture) and Fig.~\ref{fig:interpretability}(C) represents the scenario without forgetting (\eg test). 
The average mastery level of the base model ($\alpha$ = 0.0, blue line) in Fig.~\ref{fig:interpretability}(C) decreases as the student answers the questions correctly.
Therefore, the base model which is is trained under the forgetting scenarios cannot be applied to the non-forgetting scenarios.
In contrast, for the proposed model with disabled forgetting component ($add\_only$), as depicted in Fig.~\ref{fig:interpretability}(C), the average mastery level increases to one. 
This result shows that the proposed model trained for forgetting scenarios can be applied to non-forgetting scenarios as well.

% For the base model ($\alpha$ = 0.0, blue line), the average mastery level decreases as the student answers the questions correctly in Figs.~\ref{fig:interpretability}(B) and (C). 
% This result of the base model can be applicable to the scenario with forgetting like Fig.~\ref{fig:interpretability}(B), but is not applicable to the scenario without forgetting like Fig.~\ref{fig:interpretability}(C).

% The same sequence of interaction can have different learning patterns, which means different educational contexts. 
% DTKT without forgetting ($add\_only$) can be used for test and short session learning scenarios.
% For the test environment, the learning pattern of $add\_only$ can be interpreted as the DTKT estimates mastery level of the already-mastered student~\cite{learning_pattern}. 
\section{Discussion}
%DLKT is similar to temporal model in sequential aspect and similar to continual leaning in shared network side.
%DLKT는 sequential 측면에서 temporal model과 유사하며 shared network 측면에서 continual leanring과 비슷하다. 각각의 ml에서 DLKT와 비슷한 limitation이 있는지 확인과 비슷한 게 있다면 기존의 방법을 적용할 수 있는지 확인해봐야 한다.

% 왜 비교하는지를 여기에 한번 언급?

% We introduce mastery update failure and catastrophic forgetting as the side effects of DLKT.

% In this section, we argue that mastery update failure is unique 
% In this section, : temporal model, continual learning.

% We argue that limitations of DLKT caused from its unique properties. DLKT has unique properties and 

%%%% 줄이거나 언급만 해도 될 듯.  장점만 나열? 
% \textbf{Comparison to temporal model}
DLKT has similar properties to temporal model~\cite{rnn_survey} and continual learning~\cite{continual_survey}, but it has clear differences.  
We provide insights of DLKT compared to other ML problems in the Supplementary Sec. 5.

Note that we did not claim that AUROC is meaningless. AUROC is still a valid metric for evaluating performance of KT. However, it has difficulty in observing the limitations which are related to the reliability and interpretability of KT. The proposed method suggests trustworthy DLKT beyond pursuing only high AUROC.

% 두 문단 중 하나만 
% The one of the goals of proposed method is not to remove all types of forgetting, but unintended forgetting caused by side-effects of DLKT. 
% We showed that proposed method can have various learning patterns according to the value of $\alpha$. 
% Finding proper $\alpha$ for various learning environment and students with different learning rate is an interesting future work. 

% We focus on the case of correct response because wrong answer is more difficult to exploit. 
%% 틀리면서 배운다 추가 
We focus on the case that the students answer correctly, and do not consider wrong responses. 
Wrong response is very difficult to address because it can have positive meanings from the educational perspective such as near miss and reasonable attempts~\cite{mastery_learning}.
In addition, there is a study that unsuccessful attempts enhance the learning~\cite{pretesting}. 
Understanding the meaning of wrong response is challenging and can be interesting future research topic.

\bibliographystyle{aaai}
\bibliography{main}

\end{document}